\documentclass[conference]{IEEEtran}
\IEEEoverridecommandlockouts
\usepackage{cite}
\usepackage{amsmath,amssymb,amsfonts}
\usepackage{graphicx}
\usepackage{textcomp}
\usepackage{xcolor}
\usepackage{subfigure}
\usepackage{bm}
\usepackage{booktabs}
\usepackage{multirow}
\usepackage{authblk}
\usepackage{algorithm}  
\usepackage{algorithmicx}  
\usepackage{algpseudocode}
\usepackage{booktabs}
\usepackage{threeparttable}
\usepackage{balance}
\definecolor{headercolor}{gray}{0.92}
\definecolor{raprowcolor}{gray}{0.96}
\definecolor{bestcell}{RGB}{230, 245, 235} 
\usepackage{booktabs}
\usepackage{siunitx}
\usepackage{threeparttable}
\usepackage[table]{xcolor}
\usepackage{multirow}
\usepackage{pifont}
\usepackage{pgfplots}

\usepackage{comment}
\usepackage{times}  
\usepackage{helvet}  
\usepackage{courier}  
\usepackage[hyphens]{url}  
\usepackage{graphicx} 
\usepackage{multirow}
\usepackage{algorithm}
\usepackage{algpseudocode}
\usepackage{amsmath}
\usepackage[dvipsnames]{xcolor} 
\usepackage{makecell}
\usepackage{array}
\usepackage{pifont}
\usepackage{booktabs}
\usepackage{natbib}  
\usepackage{caption} 
\usepackage{booktabs}   
\usepackage{multirow}   
\usepackage{siunitx}    
\usepackage[table]{xcolor} 
\usepackage{siunitx}
\usepackage[table]{xcolor}
\usepackage{enumitem}
\usepackage{graphicx}
\usepackage{colortbl}
\usepackage{booktabs}       
\usepackage{tabularx}       
\usepackage{siunitx}        
\usepackage{xcolor}         
\usepackage{array}          
\definecolor{mygray}{gray}{0.5}
\definecolor{background_gray}{gray}{0.9}

    \def\addlegendimage{\csname pgfplots@addlegendimage\endcsname}

\definecolor{pt1min}{HTML}{1A237E}
\definecolor{pt2min}{HTML}{B71C1C}
\definecolor{pt4min}{HTML}{1B5E20}
\definecolor{wopt1min}{HTML}{01579B}
\definecolor{wopt2min}{HTML}{BF360C}
\definecolor{wopt4min}{HTML}{004D40}
\definecolor{pttte}{HTML}{E65100}
\definecolor{wopttte}{HTML}{FFD600}

\usepackage[dvipsnames]{xcolor}
\definecolor{commentcolor}{rgb}{0.45,0.45,0.45}
\definecolor{traincolor}{named}{OrangeRed}
\definecolor{inferencecolor}{named}{ForestGreen}

\DeclareMathOperator*{\argmin}{arg\,min}

\def\BibTeX{{\rm B\kern-.05em{\sc i\kern-.025em b}\kern-.08em
    T\kern-.1667em\lower.7ex\hbox{E}\kern-.125emX}}
\begin{document}



\title{Learning from History: A Retrieval-Augmented Framework for Spatiotemporal Prediction}
\def\method{\textsc{RAP}} 


\author{
	Hao Jia\textsuperscript{\rm 1},
    Penghao Zhao\textsuperscript{\rm 2},
    Hao Wu\textsuperscript{\rm 3},
    Yuan Gao\textsuperscript{\rm 3},
    Yangyu Tao\textsuperscript{\rm 2},
    Bin Cui\textsuperscript{\rm 1},\\

	\textsuperscript{\rm 1}Peking University, Beijing, China
    \textsuperscript{\rm 2}Tencent Inc., China
    \textsuperscript{\rm 3}Tsinghua University, Beijing, China \\

	haojia@stu.pku.edu.cn, hymiezhao@tencent.com,  wu-h25@mails.tsinghua.edu.cn, \\ yuangao24@mails.tsinghua.edu.cn, brucetao@tencent.com, bin.cui@pku.edu.cn \\

}

\maketitle

\begin{abstract}
Accurate and long-term spatiotemporal prediction for complex physical systems remains a fundamental challenge in scientific computing. While deep learning models, as powerful parametric approximators, have shown remarkable success, they suffer from a critical limitation: the accumulation of errors during long-term autoregressive rollouts often leads to physically implausible artifacts. This deficiency arises from their purely parametric nature, which struggles to capture the full constraints of a system's intrinsic dynamics. To address this, we introduce a novel \textbf{Retrieval-Augmented Prediction (RAP)} framework, a hybrid paradigm that synergizes the predictive power of deep networks with the grounded truth of historical data. The core philosophy of RAP is to leverage historical evolutionary exemplars as a non-parametric estimate of the system's local dynamics. For any given state, RAP efficiently retrieves the most similar historical analog from a large-scale database. The true future evolution of this analog then serves as a \textbf{reference target}. Critically, this target is not a hard constraint in the loss function but rather a powerful conditional input to a specialized dual-stream architecture. It provides strong \textbf{dynamic guidance}, steering the model's predictions towards physically viable trajectories. In extensive benchmarks across meteorology, turbulence, and fire simulation, RAP not only surpasses state-of-the-art methods but also significantly outperforms a strong \textbf{analog-only forecasting baseline}. More importantly, RAP generates predictions that are more physically realistic by effectively suppressing error divergence in long-term rollouts. Our work demonstrates that integrating non-parametric historical experience with parametric deep networks provides an effective pathway toward building the next generation of high-fidelity surrogate models for complex physical systems.\footnote{\url{https://github.com/RAP-ANAO/Retrieval-Augmented-Prediction}}

\end{abstract}

\section{Introduction}

Spatiotemporal sequence prediction is a cornerstone of many critical scientific and engineering domains, aiming to deduce the future states of complex systems from historical observations~\cite{shi2015convolutional, wang2022predrnn, wu2024earthfarsser}. In weather forecasting, accurate prediction underpins disaster mitigation and socio-economic activities~\cite{wu2025triton, bi2023accurate, gao2025oneforecast}. In computational fluid dynamics, it is crucial for revealing the intrinsic mechanisms of complex phenomena like turbulence and vortices~\cite{wu2025turb, wu2024neural, wu2024pure}. In fire simulation~\cite{feiopen, wu2024prometheus}, it provides vital decision support for evacuation strategies and emergency response. {\textit{However, these physical systems commonly exhibit highly nonlinear, multi-scale, and chaotic characteristics, which make long-term, high-fidelity prediction an exceptionally formidable scientific challenge.}}

In recent years, data-driven methods, particularly those based on deep learning models leveraging convolutions (CNNs)~\cite{gao2022simvp, tan2022simvp, wu2023pastnet}, recurrence (RNNs)~\cite{wang2022predrnn}, and attention mechanisms (Transformers)~\cite{wu2025triton, gao2022earthformer, wu2023earthfarseer}, have achieved remarkable success. These methods automatically learn complex dependencies from massive datasets, demonstrating powerful fitting capabilities. However, these models are fundamentally {\textit{purely parametric function approximators}} that learn physical laws implicitly. This lack of an explicit understanding of the system's intrinsic dynamics becomes a critical bottleneck, causing their performance to degrade sharply when facing rare events or performing long-term autoregressive rollouts~\cite{lam2023learning, wu2025triton}. Consequently, models tend to generate {physically unrealistic artifacts} and suffer from the {rapid accumulation of errors}, ultimately causing predictions to diverge and lose their value~\cite{wu2025triton}.

In stark contrast to the implicit learning of machines, human experts, such as meteorologists, often \textit{learn from history}. They search vast historical records for cases with high similarity to the current state known as {historical analogs} and use their true subsequent evolutions as a crucial reference. We argue that this classic idea contains a profound insight: every historical analog is an evolutionary exemplar validated by a real physical process, providing powerful {non-parametric information}. At a deeper level, since the evolutionary trajectory of any physical system is constrained to a low-dimensional {attractor manifold}, the path of an analog represents a real sample of the {local dynamics} on this manifold. This offers a powerful form of {dynamic guidance}, rather than a hard constraint, for the predictive model. This distinction raises a critical question: can a neural network learn to effectively \textit{fuse} this guidance with the current state, or does it merely learn to copy the analog?

Based on this philosophy, we propose a novel {Retrieval-Augmented Prediction (RAP) framework}. The RAP framework decomposes the prediction paradigm into three synergistic stages. First, in the {Retrieve} stage, for any given query, RAP employs an efficient similarity metric (e.g., Mean Squared Error) to retrieve the most similar historical analogs from a large-scale database~\cite{wang2025beamvq}. Second, in the {Augment} stage, the framework extracts the corresponding true subsequent evolutions, using them as a {reference target}. Finally, in the {Predict} stage, a specialized dual-stream network deeply fuses features from the current input and the reference target, generating a precise prediction informed by both the current state and historical precedent.

To comprehensively validate our framework, we conduct systematic evaluations on challenging benchmarks spanning {\ding{182}. Global Weather Forecasting, \ding{183}. 2D Turbulence Modeling, and \ding{184}. Fire-spread Simulation}. Our experiments are designed not only to compare RAP against state-of-the-art models but also against a strong {`analog-only` baseline}, where the prediction is simply the retrieved reference target. The results demonstrate that the RAP framework {significantly outperforms both classes of methods}. More importantly, our qualitative analysis reveals that RAP generates predictions with {more physically realistic details} and {effectively suppresses error divergence}, providing strong evidence that incorporating historical experience as an explicit form of guidance is a robust pathway to improving prediction fidelity.

In summary, the main contributions of this paper are threefold:
\begin{itemize}
    \item We propose the {RAP framework}, a novel, general, and effective retrieval-augmented paradigm that organically integrates the classic idea of historical analogs with modern deep learning models.
    \item We articulate and validate the core idea of using {historical evolutionary exemplars, retrieved from data, as a form of non-parametric dynamic guidance} to steer deep models toward more physically plausible predictions.
    \item We conduct extensive experiments demonstrating the {significant advantages} of our method, showing its superiority over both purely parametric models and strong non-learning, {analog-only baselines}, thus offering a new direction for building the next generation of high-fidelity surrogate models.
\end{itemize}
\section{Related Work}

\paragraph{Deep Learning for Spatiotemporal Prediction}

Deep learning has become the dominant paradigm for spatiotemporal prediction~\cite{shi2015convolutional, yu2018spatio, wu2024earthfarsser}, driving significant progress in diverse fields, from video prediction to climate science~\cite{wu2023pastnet}.
Early works often employed architectures based on Convolutional Neural Networks (CNNs)~\cite{he2016deep}, such as U-Net~\cite{ronneberger2015u}, which excel at capturing local spatial features but are limited in their ability to explicitly model temporal dependencies.
To better handle time-series data, researchers integrated Recurrent Neural Networks (RNNs) with CNNs, leading to models like ConvLSTM~\cite{shi2015convolutional} and PredRNN~\cite{wang2022predrnn}.
These models leverage recurrent state units to propagate temporal information, improving the modeling of long-range dependencies.
However, their sequential nature restricts parallel computation and they still face challenges like vanishing gradients in very long sequences.
More recently, inspired by the success of Transformers in natural language processing and computer vision, a series of attention-based models, including FourCastNet~\cite{kurth2023fourcastnet}, Pangu-Weather~\cite{bi2023accurate}, and Triton~\cite{wu2025triton}, have achieved breakthrough performance in the prediction of physical systems.
By leveraging global attention mechanisms, they effectively capture long-range spatial dependencies and have established the current state-of-the-art.
However, all of these models adhere to a purely parametric paradigm, compressing the complex dynamics of an entire physical system into a finite set of fixed network weights.
This "one-size-fits-all" approach presents a fundamental challenge to their ability to generalize to rare events or maintain stability over long-term rollouts.
In essence, despite their growing capabilities, existing deep learning models lack a mechanism to explicitly leverage verified, real-world examples of evolution from history as guidance. Our work aims to fill this key gap.

\paragraph{Retrieval-Augmented Models}

The "retrieve-and-augment" paradigm has recently demonstrated immense potential in other machine learning domains.
The most prominent example is in Natural Language Processing (NLP), where Retrieval-Augmented Generation (RAG)~\cite{zhao2024retrieval} models enhance the outputs of language models by retrieving relevant text snippets from an external knowledge base. This approach effectively mitigates factual errors and "hallucinations" in generated text.
Similarly, in the field of Computer Vision~\cite{wangnuwadynamics}, exemplar-based methods for tasks like image inpainting and synthesis generate high-quality results by borrowing realistic textures and structures from reference images.
Inspired by these successes, our work is the first to systematically introduce the retrieval-augmented philosophy to the prediction of complex physical processes.
Unlike RAG, which retrieves discrete text fragments, our {RAP framework} is designed to retrieve and utilize high-dimensional, continuous spatiotemporal fields.
We explore how these retrieved physical exemplars can serve as a {dynamical constraint} to guide and regularize the behavior of a deep predictive model.

\paragraph{Analog Forecasting in Scientific Computing}

It is noteworthy that the idea of learning from history has a long and storied tradition in scientific computing, particularly in meteorology~\cite{liu2024retrieval}, where it is known as {Analog Forecasting}.
Since the pioneering work of Lorenz, forecasters have assisted in their predictions by identifying analogs, past atmospheric states that most closely resemble the current one.
However, traditional analog methods typically rely on hand-crafted similarity metrics and are difficult to integrate seamlessly and end-to-end with modern deep learning models~\cite{wu2023LSM, wu2021autoformer, wu2023earthfarseer}.
They often exist as a post-processing step or as an independent source of reference information.
Our RAP framework can be viewed as a modern, end-to-end realization of this classic philosophy.
We integrate the retrieval process directly into a deep learning pipeline, enabling the model not only to leverage the retrieved information but also to learn \textit{how best} to utilize it during training.

\section{Methodology}

\begin{table}[t]
\centering
\caption{Summary of key notations.}
\label{tab:notations_single_column}
\small
\renewcommand{\arraystretch}{1.2}
\begin{tabularx}{\columnwidth}{@{} c X @{}}
\toprule
{Symbol} & {Description} \\ 
\midrule
$\mathbf{X}$, $\mathbf{Y}$ & Input and output sequences \\
$\hat{\mathbf{Y}}$ & Predicted future sequence \\
$T_{in}$, $T_{out}$ & Input/output timesteps \\
$C, H, W$ & Channels, height, and width of a field \\
$\mathcal{F}$, $\mathcal{F}'$ & Standard and RAP-enhanced models \\
$\mathbf{X}_{\text{query}}$ & Query input sequence \\
$\mathbf{Y}_{\text{ref}}$ & Reference Target from a historical analog \\
$\mathbf{Y}_{\text{gt}}$ & Ground-truth future for $\mathbf{X}_{\text{query}}$ \\
$\mathcal{D}$ & Historical database of $N$ pairs \\
$\mathcal{S}(\cdot, \cdot)$ & Similarity metric for retrieval \\
$\mathcal{E}_{\text{query}}, \mathcal{E}_{\text{ref}}$ & Query and reference encoders \\
$\theta_{\text{query}}, \theta_{\text{ref}}$ & Parameters of query/reference encoders \\
$\mathcal{L}_{\text{total}}$ & Total training loss function \\
$\Theta$ & All learnable parameters in the RAP model \\ 
\bottomrule
\end{tabularx}
\end{table}
In this section, we provide a comprehensive exposition of our proposed Retrieval-Augmented Prediction (RAP) meta-framework. This framework is engineered to fundamentally address the critical limitations of purely parametric models—namely, their propensity for error accumulation and generation of physically implausible artifacts during long-term forecasting. The central tenet of RAP is to anchor the predictions of a deep neural network in the reality of historical data by integrating verified evolutionary exemplars as an explicit conditional input. This retrieved information does not act as a hard optimization constraint, but rather as a powerful form of {non-parametric dynamic guidance}, steering the model towards physically consistent and stable trajectories. We begin by formalizing the spatiotemporal prediction task, then meticulously detail each component of the RAP framework. A summary of the key notations is provided in Table~\ref{tab:notations_single_column}.
\subsection{Problem Formulation}

From a probabilistic perspective, the task of spatiotemporal sequence prediction is to model the conditional probability distribution $p(\mathbf{Y} | \mathbf{X})$ of a future state sequence $\mathbf{Y} \in \mathbb{R}^{T_{out} \times C \times H \times W}$, given a sequence of past observations $\mathbf{X} \in \mathbb{R}^{T_{in} \times C \times H \times W}$. A deep learning model, parameterized by $\Theta$, aims to learn this distribution, denoted as $p_{\Theta}(\mathbf{Y} | \mathbf{X})$.

The standard training paradigm for such models is Maximum Likelihood Estimation (MLE). Given a dataset of observed trajectories $\mathcal{D} = \{(\mathbf{X}_i, \mathbf{Y}_i)\}_{i=1}^{N}$, the objective is to find the optimal parameters $\Theta^*$ that maximize the log-likelihood of observing the ground-truth future sequences:
\begin{equation}
    \Theta^* = \arg\max_{\Theta} \sum_{i=1}^{N} \log p_{\Theta}(\mathbf{Y}_i | \mathbf{X}_i)
\end{equation}
Typically, this is achieved by assuming the conditional distribution follows a simple parametric form, such as a Gaussian distribution with a mean predicted by a deterministic function $\mathcal{F}_{\Theta}(\mathbf{X})$ and a fixed variance. Under this assumption, maximizing the log-likelihood is equivalent to minimizing the Mean Squared Error (MSE) between the prediction $\hat{\mathbf{Y}} = \mathcal{F}_{\Theta}(\mathbf{X})$ and the ground truth $\mathbf{Y}$. This leads to the conventional objective of learning a monolithic mapping:
\begin{equation}
    \hat{\mathbf{Y}} = \mathcal{F}(\mathbf{X})
\end{equation}

However, we argue that for complex physical systems, this monolithic conditioning on $\mathbf{X}$ is insufficient to capture the intricate dynamics, often leading to physically implausible predictions. The core philosophy of our Retrieval-Augmented Prediction (RAP) framework is to enrich this conditioning with dynamically-relevant, non-parametric information. Specifically, we introduce an additional conditioning variable: a retrieved {historical analog}.

This reframes the problem. Instead of modeling $p(\mathbf{Y} | \mathbf{X})$, we aim to model a more informed conditional distribution, $p(\mathbf{Y} | \mathbf{X}_{\text{query}}, \mathbf{Y}_{\text{ref}})$, where $\mathbf{Y}_{\text{ref}}$ is the true future evolution of the historical analog most similar to the current query, $\mathbf{X}_{\text{query}}$. Our RAP model, $\mathcal{F}'$, is thus designed to approximate the mean of this richer distribution. The learning objective, now viewed through the lens of conditional MLE, becomes finding a mapping that generates the most likely future, given both the current state and a historical precedent:
\begin{equation}
    \hat{\mathbf{Y}} = \mathcal{F}'(\mathbf{X}_{\text{query}}, \mathbf{Y}_{\text{ref}})
\end{equation}
By incorporating $\mathbf{Y}_{\text{ref}}$, we provide the model with a strong inductive bias towards physically valid trajectories, effectively regularizing the solution space and guiding the prediction towards a more robust and accurate outcome.

\subsection{The RAP Framework}

\begin{figure*}[h!]
    \centering
    \includegraphics[width=\textwidth]{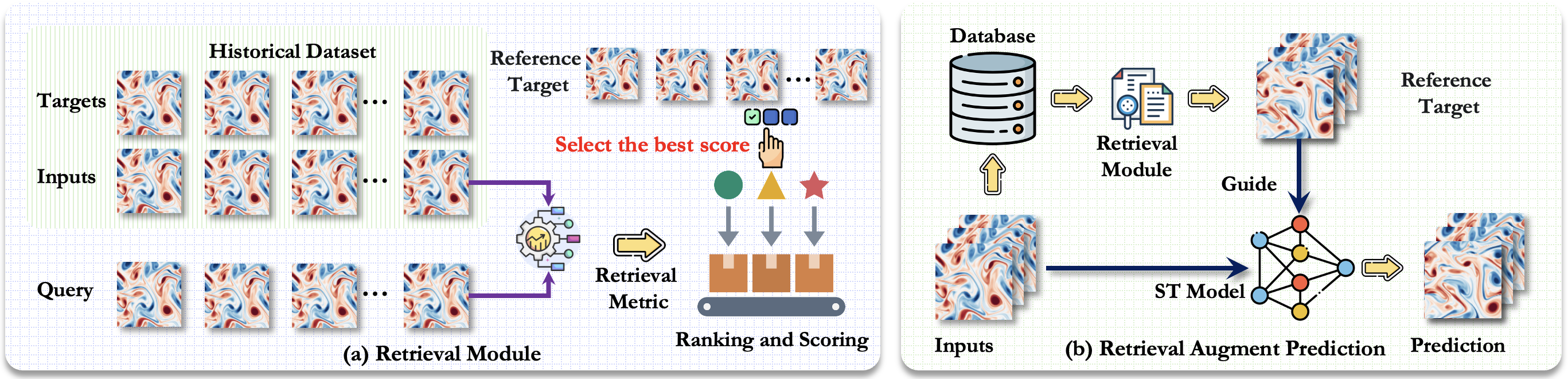}
    \caption{
    {An overview of the Retrieval-Augmented Prediction (RAP) framework.}
    Given a \texttt{Query Input}, a similarity search is performed on a \texttt{Historical Database} to find the best analog, whose future evolution is designated as the \texttt{Reference Target}. 
    This reference is then used in two ways: 
    {(a)} directly as the prediction for the {analog-only baseline}, and 
    {(b)} as a conditional input (\texttt{dynamic guidance}) to our {Dual-Stream Fusion Model}, which also takes the original query to generate the final \texttt{RAP Prediction}. 
    Both predictions are evaluated against the \texttt{Ground Truth Future}.
}
    \label{fig:rap_framework}
\end{figure*}

RAP is a general, three-stage meta-framework: \textit{Retrieve}, \textit{Augment}, and \textit{Predict}. As illustrated in Figure~\ref{fig:rap_framework}, for any given query sequence $\mathbf{X}_{\text{query}}$, the framework first executes a retrieval operation to find a pertinent historical analog. This analog's true future is then used to augment a deep predictive model, which synthesizes both sources of information to render a final, high-fidelity prediction.
\subsubsection{Stage 1: Historical Analog Retrieval}

The primary objective of this stage is to locate the "nearest neighbor" of the current query $\mathbf{X}_{\text{query}}$ within the system's high-dimensional dynamical manifold. This is achieved by searching a large-scale historical database, $\mathcal{D} = \{(\mathbf{X}_i, \mathbf{Y}_i)\}_{i=1}^{N}$, where each entry is a complete, historically observed trajectory.

\paragraph{Historical Analog Database.} The power of our framework stems from this database, $\mathcal{D}$. It is not merely a collection of data points, but a repository of $N$ evolutionary exemplars. Each pair $(\mathbf{X}_i, \mathbf{Y}_i)$ represents a complete cause-and-effect sequence that has been validated by the immutable laws of a real physical process. This database serves as a non-parametric memory of the system's dynamics, capturing a vast array of behaviors, including rare and extreme events, that a purely parametric model might struggle to learn or generalize to.

\paragraph{Similarity Metric and Retrieval.} To find the most relevant analog, we define a similarity metric $\mathcal{S}: \mathbb{R}^{T_{in} \times C \times H \times W} \times \mathbb{R}^{T_{in} \times C \times H \times W} \to \mathbb{R}^+$. For this study, we deliberately employ a foundational and computationally efficient metric, the Mean Squared Error (MSE), to establish a strong and reproducible baseline. The metric is expressed as the normalized squared Frobenius norm of the difference between two sequences:
\begin{equation}
    \mathcal{S}(\mathbf{X}_a, \mathbf{X}_b) = \frac{1}{T_{in} \cdot C \cdot H \cdot W} \sum_{t=1}^{T_{in}} \| \mathbf{x}_{a,t} - \mathbf{x}_{b,t} \|_F^2
\end{equation}
The retrieval process is then a minimization problem over the entire database to find the index $k$ of the best-matching historical input:
\begin{equation}
    k = \arg\min_{i \in \{1, \dots, N\}} \mathcal{S}(\mathbf{X}_{\text{query}}, \mathbf{X}_i)
\end{equation}
This operation yields the {historical analog}, $\mathbf{X}_{\text{analog}} = \mathbf{X}_k$. The subsequent, ground-truth evolution corresponding to this analog, $\mathbf{Y}_k$, is then designated as the {Reference Target}, $\mathbf{Y}_{\text{ref}}$.

\paragraph{Justification and Discussion on the Choice of Metric.} We explicitly acknowledge a critical point raised by the scientific community: a pixel-wise metric like MSE has inherent limitations. It is, by definition, sensitive to spatial translations, rotations, or phase shifts. For instance, two atmospheric states representing the same weather pattern (e.g., a vortex) but shifted by a few grid cells would register a large MSE, despite being dynamically very similar points on the system's attractor. Similarly, two periodic signals, such as $\sin(t)$ and $\sin(t+\phi)$, would have a non-zero MSE despite identical dynamics. While more sophisticated, potentially invariant metrics (e.g., based on Fourier transforms, optimal transport, or learned perceptual features) are a crucial and exciting avenue for future research, our choice of MSE is deliberate for this foundational work. It is simple, universally understood, computationally inexpensive, and, as our results will show, surprisingly effective. By demonstrating significant gains even with this basic metric, we establish a robust lower bound on the potential of the RAP paradigm, isolating the contribution of the neural fusion mechanism itself.

\subsubsection{Stages 2 \& 3: Augmentation and Prediction}

This phase concerns the core of our contribution: the principled integration of the non-parametric information embodied in $\mathbf{Y}_{\text{ref}}$ into a parametric prediction model.

\paragraph{Framework Generality and Model-Agnosticism.} A key strength of RAP is its nature as a {model-agnostic} paradigm, not a monolithic architecture. It can be applied to augment any state-of-the-art spatiotemporal model, whether based on CNNs, RNNs, or Transformers. This versatility elevates a standard single-input model, $\mathcal{F}(\mathbf{X})$, into a more powerful, conditionally-guided dual-input model, $\mathcal{F}'(\mathbf{X}_{\text{query}}, \mathbf{Y}_{\text{ref}})$. This highlights that our contribution is a general strategy for improving spatiotemporal forecasting, granting the RAP framework immense flexibility and scalability.

\paragraph{Model Instantiation: A Dual-Stream Spatiotemporal Network.} To provide a concrete and effective demonstration of the RAP paradigm, we instantiate a specialized dual-stream neural network. This network is meticulously designed with three primary components: a dual-stream encoder, a feature fusion module, and a hierarchical decoder. We now elaborate on the architecture and the rationale behind its design, moving beyond a simple list of components.

The network architecture begins with a {Dual-Stream Encoder}. This design choice is fundamental to the RAP philosophy. We employ two parallel encoder pathways: a query encoder, $\mathcal{E}_{\text{query}}$, parameterized by $\theta_{\text{query}}$, and a reference encoder, $\mathcal{E}_{\text{ref}}$, parameterized by $\theta_{\text{ref}}$. A critical design decision is that these two encoders, while sharing an identical architectural blueprint (e.g., a U-Net like structure), possess {independent and unshared parameters}. The rationale for this is to allow for specialized feature extraction tailored to the distinct nature of their respective inputs. The query encoder, $\mathcal{E}_{\text{query}}$, processes $\mathbf{X}_{\text{query}}$, which represents the precise, but incomplete, current state of the system; its task is to learn a sensitive representation of initial conditions. In contrast, the reference encoder, $\mathcal{E}_{\text{ref}}$, processes $\mathbf{Y}_{\text{ref}}$, which is a complete, dynamically consistent, ground-truth future trajectory; its task is to learn to extract robust, archetypal patterns of physical evolution. Forcing these two distinct tasks through a single, shared-parameter encoder would inevitably lead to a compromised, less effective representation. Each encoder, $\mathcal{E}$, maps its input sequence $\mathbf{Z}$ to two outputs: a deep, low-resolution semantic representation, $\mathbf{h}_{\text{latent}}$, capturing the global context, and a set of multi-scale, high-resolution feature maps, $\mathcal{H}_{\text{skip}} = \{\mathbf{h}_{\text{skip}}^{(l)}\}_{l=1}^{L-1}$, destined for skip connections. Mathematically:
\begin{equation}
    (\mathbf{h}_{\text{latent}}, \mathcal{H}_{\text{skip}}) = \mathcal{E}(\mathbf{Z}; \theta)
\end{equation}
This process yields two distinct sets of learned features, one for the query and one for the reference:
\begin{align}
    (\mathbf{h}_{\text{latent}}^{\text{query}}, \mathcal{H}_{\text{skip}}^{\text{query}}) &= \mathcal{E}_{\text{query}}(\mathbf{X}_{\text{query}}; \theta_{\text{query}}) \\
    (\mathbf{h}_{\text{latent}}^{\text{ref}}, \mathcal{H}_{\text{skip}}^{\text{ref}}) &= \mathcal{E}_{\text{ref}}(\mathbf{Y}_{\text{ref}}; \theta_{\text{ref}})
\end{align}

Following encoding, the process moves to {Feature Fusion and Hierarchical Decoding}. The deep semantic representations from both streams are first integrated. In our instantiation, we employ a simple yet effective channel-wise concatenation to form a fused latent representation, $\mathbf{h}_{\text{fused}} = \text{Concat}(\mathbf{h}_{\text{latent}}^{\text{query}}, \mathbf{h}_{\text{latent}}^{\text{ref}})$. This fused vector now contains a rich, combined understanding of both the system's current specific state and a template for its plausible future evolution. This fused representation then seeds a {hierarchical decoder}, $\mathcal{G}$, which is composed of a series of $L$ upsampling blocks, $\{\mathcal{G}_l\}_{l=1}^L$. The decoder's task is to progressively upsample this low-resolution latent state back to the full spatiotemporal resolution of the desired prediction. To preserve the fine-grained physical details crucial for high-fidelity forecasting, the decoder incorporates skip connections at each of its hierarchical stages. Crucially, it receives skip connections from \textit{both} of the original encoders. At a given decoding stage $l$, the upsampled features from the previous stage, $\mathbf{z}_{l-1}$, are combined with the corresponding scale-matched skip features from both the query, $\mathbf{h}_{\text{skip}}^{\text{query},(L-l)}$, and the reference, $\mathbf{h}_{\text{skip}}^{\text{ref},(L-l)}$. This multi-scale fusion ensures that the final prediction is constructed using both the precise details of the initial state (from the query) and the plausible textural and structural patterns of a real evolution (from the reference). The process can be expressed as follows, where $\mathbf{z}_0 = \mathbf{h}_{\text{fused}}$:
\begin{equation}
\begin{split}
    \mathbf{z}_{l} = \mathcal{G}_l \bigg( & \text{UpSample}(\mathbf{z}_{l-1}), \\
                                     & \text{Concat}\big(\mathbf{h}_{\text{skip}}^{\text{query},(L-l)}, \mathbf{h}_{\text{skip}}^{\text{ref},(L-l)}\big) \bigg), \\
                                     & \qquad \qquad \text{for } l=1, \dots, L
\end{split}
\end{equation}
The final output of the last decoder block, $\hat{\mathbf{Y}} = \mathbf{z}_L$, constitutes the model's definitive prediction.

\subsection{Optimization Objective}

The training of the entire RAP model is supervised by minimizing the discrepancy between the model's prediction, $\hat{\mathbf{Y}}$, and the actual observed {ground-truth} future, $\mathbf{Y}_{\text{gt}}$.

\paragraph{The Role of the Reference Target as Dynamic Guidance.} It is of paramount importance to clarify a point of potential confusion that is central to our contribution. The reference target, $\mathbf{Y}_{\text{ref}}$, is {exclusively used as a conditional input during the forward pass and is completely absent from the loss computation}. This design is a deliberate and principled choice that directly addresses the semantic difference between a "constraint" and "guidance." If $\mathbf{Y}_{\text{ref}}$ were part of the loss function, the network would be incentivized to simply learn an identity mapping or a trivial copy mechanism. By excluding it from the loss, we force the network to learn a more sophisticated task: it must treat $\mathbf{Y}_{\text{ref}}$ as a source of information about plausible physical dynamics and learn how to \textit{synthesize} this information with the specific initial conditions provided by $\mathbf{X}_{\text{query}}$ to produce an accurate prediction of the true, unseen future, $\mathbf{Y}_{\text{gt}}$. Therefore, $\mathbf{Y}_{\text{ref}}$ does not function as a "constraint" in the strict optimization-theoretic sense. Instead, it acts as a powerful {non-parametric inductive bias} or a form of {dynamic guidance} that regularizes the vast solution space of the deep network, pushing it away from physically unrealistic trajectories.

\paragraph{Loss Function.} To train the model, we employ a composite loss function that combines the Mean Absolute Error (L1 loss) and the Mean Squared Error (L2 loss). This combination is a standard practice that balances the L1 loss's robustness to outliers with the L2 loss's sensitivity to large errors, promoting predictions that are both accurate in pixel values and structurally smooth. The total loss, $\mathcal{L}_{\text{total}}$, is a weighted sum:
\begin{equation}
    \mathcal{L}_{\text{total}}(\hat{\mathbf{Y}}, \mathbf{Y}_{\text{gt}}) = \lambda_1 \mathcal{L}_{1}(\hat{\mathbf{Y}}, \mathbf{Y}_{\text{gt}}) + \lambda_2 \mathcal{L}_{\text{MSE}}(\hat{\mathbf{Y}}, \mathbf{Y}_{\text{gt}})
\end{equation}
To ensure reproducibility and address inquiries about hyperparameter selection, we state that across all diverse datasets and backbone models in our experiments, the balancing weights were held constant at $\lambda_1 = 1.0$ and $\lambda_2 = 1.0$. This simple and consistent setting proved to be robust and effective, obviating the need for task-specific tuning. The individual, fully-normalized loss terms are defined as:
\begin{align}
    \mathcal{L}_{1} &= \frac{1}{T_{out}} \sum_{t=1}^{T_{out}} \frac{1}{CWH} \sum_{c,h,w} |\hat{y}_{t,c,h,w} - y_{\text{gt},t,c,h,w}| \\
    \mathcal{L}_{\text{MSE}} &= \frac{1}{T_{out}} \sum_{t=1}^{T_{out}} \frac{1}{CWH} \sum_{c,h,w} (\hat{y}_{t,c,h,w} - y_{\text{gt},t,c,h,w})^2
\end{align}
The entire set of learnable parameters in the model, $\Theta = \{\theta_{\text{query}}, \theta_{\text{ref}}, \theta_{\text{dec}}\}$, is optimized end-to-end with respect to this total loss using backpropagation and the Adam optimizer.

\paragraph{In summary} Our proposed RAP framework operates by first retrieving a pertinent historical analog from a large-scale database based on its similarity to the current query. The true subsequent evolution of this analog then serves as a non-parametric, dynamic guide to augment a deep learning prediction model. This model synthesizes information from both the current state and the historical reference to generate a final, high-fidelity prediction. To formalize this operational flow, we present the core mechanics of the RAP framework in Algorithm~\ref{alg:rap_framework_concise}, which details the key procedures for both training and inference.
\begin{algorithm}[H]
\caption{The Concise RAP Framework}
\label{alg:rap_framework_concise}
\begin{algorithmic}[1]
\State \textbf{Define:} Dual-stream model $\mathcal{F}'(X_{\text{query}}, Y_{\text{ref}}; \Theta)$, Similarity metric $S(\cdot, \cdot)$
\Statex 
\Procedure{TrainStep}{$X_{\text{query}}, Y_{\text{gt}}$}
    \State $k \gets \argmin_{i} S(X_{\text{query}}, X_i)$ for $(X_i, Y_i) \in \mathcal{D}_{\text{hist}}$ \Comment{1. Retrieve analog}
    \State $\hat{Y} \gets \mathcal{F}'(X_{\text{query}}, Y_k; \Theta)$ \Comment{2. Augment and Predict}
    \State $\mathcal{L} \gets \text{Loss}(\hat{Y}, Y_{\text{gt}})$ \Comment{3. Compute loss against ground truth}
    \State Update $\Theta$ using $\nabla_{\Theta} \mathcal{L}$ \Comment{4. Optimize parameters}
\EndProcedure
\Statex
\Procedure{Inference}{$X_{\text{new}}$}
    \State $k \gets \argmin_{i} S(X_{\text{new}}, X_i)$ for $(X_i, Y_i) \in \mathcal{D}_{\text{hist}}$ \Comment{1. Retrieve analog}
    \State $\hat{Y}_{\text{new}} \gets \mathcal{F}'(X_{\text{new}}, Y_k; \Theta_{\text{trained}})$ \Comment{2. Generate prediction}
    \State \Return $\hat{Y}_{\text{new}}$
\EndProcedure
\end{algorithmic}
\end{algorithm} 
\begin{table*}[h] 
\centering
\begin{threeparttable} 
\caption{A comprehensive summary of the key specifications for the three benchmark datasets employed in our evaluation. The table details the number of samples in the training, validation, and test sets. It also specifies the tensor shapes for both input and output data, following the format.}
\label{tab:dataset_specs}
\small 
\renewcommand{\arraystretch}{1.2}
\definecolor{headercolor}{gray}{0.92}
\sisetup{group-separator={,}} 
\begin{tabularx}{\textwidth}{@{} l *{3}{>{\centering\arraybackslash}X} @{}}
\toprule
\rowcolor{headercolor}
\textbf{Specification} & \textbf{ERA5} & \textbf{2D Turbulence} & \textbf{Prometheus} \\
\midrule
Train Set Size & \num{4485} & \num{101376} & \num{13545} \\
Validation Set Size & \num{460} & \num{12672} & \num{1505} \\
Test Set Size & \num{460} & \num{12672} & \num{1505} \\
Input Tensor Shape & (1, 69, 180, 360) & (1, 1, 128, 128) & (10, 2, 32, 480) \\
Output Tensor Shape & (1, 69, 180, 360) & (1, 1, 128, 128) & (10, 2, 32, 480) \\
\bottomrule
\end{tabularx}
\end{threeparttable}
\end{table*}

\section{Experiments}
\label{sec:experiments}

\subsection{Summary of Experimental Findings}

Based on our comprehensive experiments, we have validated the following core contributions of the Retrieval-Augmented Prediction (RAP) framework:

\begin{itemize}
    \item \textbf{RQ1: \textit{Effectiveness and Generality}:} Our experiments consistently demonstrate RAP's effectiveness and generality. Across a wide spectrum of baseline models spanning CNN, Transformer, and Fourier/Operator-based architectures and diverse, challenging physical domains such as weather forecasting, turbulence modeling, and fire simulation, the RAP framework yields significant and systematic improvements in prediction accuracy (Table~\ref{tab:main_results}, Figure~\ref{fig:relative_improvement}). This universal enhancement underscores that leveraging historical analogs as explicit dynamical guidance is a fundamentally beneficial and model-agnostic paradigm.

    \item \textbf{RQ2: \textit{Physical Fidelity and Stability}:} RAP significantly enhances the physical fidelity and long-term stability of predictions. Beyond superior quantitative metrics, qualitative results reveal that RAP-enhanced models effectively suppress error accumulation during autoregressive rollouts. They successfully preserve critical, high-frequency physical structures, such as sharp vortices in turbulence and intricate flame fronts in fire simulations which are often blurred or lost by baseline models due to numerical dissipation (Figure~\ref{fig:qualitative_comparison}, Figure~\ref{fig:prometheus_qualitative}).

    \item \textbf{RQ3: \textit{Robustness and Design Principles}:} The architectural design of the RAP framework is crucial for its success. Our extensive ablation studies validate the superiority of our proposed dual-stream architecture over naive concatenation methods, proving that \textit{how} historical information is integrated is more critical than its mere presence (Table~\ref{tab:ablation_fusion_necessity}, Figure~\ref{fig:qualitative_fusion_necessity}). Furthermore, our findings suggest that an asymmetric design, utilizing a lightweight convolutional encoder for the reference stream, is optimal for extracting robust dynamic patterns from the ground-truth reference target.

    \item \textbf{RQ4: \textit{Scalability and Training Efficiency}:} RAP demonstrates remarkable scalability and offers a path toward greater training efficiency for large-scale models (LSMs). By applying RAP to a 0.6B parameter Triton model, we show that leveraging a large, static historical database can substantially mitigate performance degradation from reduced training data. Specifically, RAP recovered approximately \textbf{81\%} of the performance gap, showcasing its potential to partially substitute expensive, full-dataset training with efficient, retrieval-based knowledge integration, thereby reducing computational costs (Table~\ref{tab:scalability}).
\end{itemize}

\subsection{Experimental Setup}

\subsubsection{Datasets and Tasks}
Our empirical evaluation is conducted on three challenging and widely-recognized public benchmarks, each representing a distinct and complex domain in scientific computing. Key statistics and data specifications for these datasets are summarized in Table~\ref{tab:dataset_specs}.

\begin{itemize}
    \setlength\itemsep{0.5em}
    \item[\ding{182}] \textbf{\textit{Weather Forecasting (ERA5)}}: We utilize the ERA5 reanalysis dataset~\cite{hersbach2020era5}, the de facto gold standard for weather and climate science. The task is to forecast the future evolution of 69 atmospheric variables. The highly chaotic nature, multi-scale dynamics, and vast data scale of this benchmark present a formidable challenge for long-term prediction fidelity.
    
    \item[\ding{183}] \textbf{\textit{2D Turbulence Modeling}}: This task involves predicting the vorticity field evolution in a 2D incompressible fluid governed by the Navier-Stokes equations. Its complex, nonlinear energy cascade from large to small scales serves as an ideal testbed for evaluating a model's ability to capture fine-grained physical dynamics and preserve high-frequency details.
    
    \item[\ding{184}] \textbf{\textit{Fire Spread Simulation (Prometheus)}}: We use a multi-physics simulation dataset for wildfire spread, which involves the coupled dynamics of temperature and fuel density. The primary challenge lies in accurately modeling highly nonlinear processes such as combustion and heat convection, which demands exceptional physical consistency from the predictive model.
\end{itemize}

For all tasks, we construct a large-scale {Historical Database} for retrieval. Unless otherwise specified, we partition the data such that trajectories from 1979--2017 are used for training and retrieval, while those from 2018--2021 are reserved for testing.

\subsubsection{Baseline Models}
To validate the effectiveness and generality of our proposed RAP framework, we select a comprehensive suite of representative spatiotemporal prediction models as baselines. These models span the three dominant architectural paradigms in the field, allowing us to assess RAP's compatibility and performance gains across diverse methodological approaches. The selected baselines are as follows:

\begin{itemize}
    \setlength\itemsep{0em}
    \item \textbf{\textit{CNN-based Models}}: We include {SimVP}~\cite{gao2022simvp}, a simple yet effective model for video prediction; {UNet}~\cite{ronneberger2015u}, a cornerstone architecture in scientific computing known for its hierarchical feature representation; and {ResNet}~\cite{he2016deep}, a classic deep residual network adapted for spatiotemporal tasks.
    
    \item \textbf{\textit{Transformer-based Models}}: This category features {DiT}~\cite{paparelladiffusion}, a Diffusion Transformer that has shown promise in generative modeling for physical systems, and {Triton}~\cite{wu2025triton}, a state-of-the-art model designed for high-fidelity, long-term Earth system forecasting.
    
    \item \textbf{\textit{Fourier/Operator-based Models}}: We evaluate {CNO}~\cite{raonic2023convolutional}, a Convolutional Neural Operator that combines the strengths of CNNs and operator learning, and {MGNO}~\cite{he2024mgno}, a multigrid neural operator designed for efficient parameterization of linear operators.
\end{itemize}

For each selected model, we conduct a rigorous and fair comparison between its original implementation, denoted as the {Baseline}, and its RAP-enhanced counterpart, denoted as {+ RAP}. This allows for a direct and unambiguous assessment of the performance improvements attributable to our framework.

\subsubsection{Evaluation Metrics}
To provide a comprehensive and multi-faceted assessment of prediction performance, we employ a set of standard evaluation metrics. Let $\hat{Y}$ denote the predicted tensor and $Y$ represent the ground truth tensor, both with dimensions $T_{out} \times C \times H \times W$. The metrics are defined as follows:

\begin{itemize}
    \setlength\itemsep{0.5em}
    \item \textbf{Mean Squared Error (MSE, $\downarrow$)}: Measures the average squared difference between the predicted and actual values at the pixel level. It is highly sensitive to large errors.
    \begin{equation*}
        \text{MSE} = \frac{1}{T_{out} \cdot C \cdot H \cdot W} \sum_{t,c,h,w} (\hat{Y}_{t,c,h,w} - Y_{t,c,h,w})^2
    \end{equation*}

    \item \textbf{Mean Absolute Error (MAE, $\downarrow$)}: Computes the average absolute difference, providing a more robust measure of pixel-level discrepancies that is less sensitive to outliers than MSE.
    \begin{equation*}
        \text{MAE} = \frac{1}{T_{out} \cdot C \cdot H \cdot W} \sum_{t,c,h,w} |\hat{Y}_{t,c,h,w} - Y_{t,c,h,w}|
    \end{equation*}

    \item \textbf{Peak Signal-to-Noise Ratio (PSNR, $\uparrow$)}: A logarithmic metric derived from MSE that quantifies the ratio between the maximum possible power of a signal and the power of corrupting noise, commonly used to assess reconstruction quality.
    \begin{equation*}
        \text{PSNR} = 10 \cdot \log_{10} \left( \frac{\text{MAX}_I^2}{\text{MSE}} \right)
    \end{equation*}
    where $\text{MAX}_I$ is the maximum possible pixel value of the data (e.g., 1 for data normalized to $[0, 1]$).

    \item \textbf{Structural Similarity Index (SSIM, $\uparrow$)}: Measures the perceptual similarity between two images by comparing their local patterns of luminance, contrast, and structure, which often aligns better with human visual assessment. For two corresponding image windows $x$ and $y$, it is defined as:
    \begin{equation*}
        \text{SSIM}(x, y) = \frac{(2\mu_x\mu_y + c_1)(2\sigma_{xy} + c_2)}{(\mu_x^2 + \mu_y^2 + c_1)(\sigma_x^2 + \sigma_y^2 + c_2)}
    \end{equation*}
    where $\mu$, $\sigma^2$, and $\sigma_{xy}$ are the local means, variances, and covariance, respectively. $c_1$ and $c_2$ are stabilization constants. The final score is the mean SSIM value over all windows.
\end{itemize}

All metrics are reported as averages over the entire prediction horizon ($T_{out}$).

\subsubsection{Implementation Details}
\paragraph{Training} All models are implemented in PyTorch and trained on NVIDIA A100 GPUs. We use the Adam optimizer with a learning rate of $1 \times 10^{-4}$ and a cosine annealing scheduler. The total loss is a weighted sum of L1 and MSE losses, i.e., $L_{\text{total}} = \lambda_1 \mathcal{L}_1 + \lambda_2 \mathcal{L}_{\text{MSE}}$.

\paragraph{RAP Configuration} The historical analog retrieval is based on {Mean Squared Error (MSE)} between the query input $X_{\text{query}}$ and candidates from the historical database. We select the single best match ($k=1$), and its corresponding ground-truth future $Y_k$ serves as the {Reference Target} $Y_{\text{ref}}$. To manage the large-scale ERA5 database efficiently, we build a compact yet representative retrieval library by sampling the historical data with an interval of 3 time steps.

\paragraph{Dual-Stream Architecture} Our RAP-enhanced models employ a dual-stream architecture. The {Query Stream} uses the same encoder as the baseline model for a fair comparison. The {Reference Stream}, by default, utilizes a lightweight {Encoder} to process $Y_{\text{ref}}$. The parameters of the two stream encoders are kept separate and are not shared.

\begin{table*}[t]
\centering
\begin{threeparttable} 
\caption{Quantitative comparison of baseline models versus their RAP-enhanced counterparts across diverse benchmarks.}
\label{tab:main_results}

\renewcommand{\arraystretch}{1.3}
\small
\setlength{\tabcolsep}{4pt}

\definecolor{headercolor}{gray}{0.92}
\definecolor{raprowcolor}{gray}{0.96}
\definecolor{bestcell}{RGB}{230, 245, 235}

\sisetup{detect-weight, mode=text}

\begin{tabular}{@{} l l S[table-format=1.4] S[table-format=1.4] S[table-format=2.2] S[table-format=1.4]
                      S[table-format=1.4] S[table-format=1.4] S[table-format=2.2] S[table-format=1.4]
                      S[table-format=1.4] S[table-format=1.4] S[table-format=2.2] S[table-format=1.4] @{}}
\toprule
\multirow{2}{*}{\textbf{Model}} & \multirow{2}{*}{\textbf{Variant}} & \multicolumn{4}{c}{\textbf{Weatherbench2}} & \multicolumn{4}{c}{\textbf{Prometheus}} & \multicolumn{4}{c}{\textbf{Turbulent}} \\
\cmidrule(lr){3-6} \cmidrule(lr){7-10} \cmidrule(lr){11-14}
& & {MSE $\downarrow$} & {MAE $\downarrow$} & {PSNR $\uparrow$} & {SSIM $\uparrow$} & {MSE $\downarrow$} & {MAE $\downarrow$} & {PSNR $\uparrow$} & {SSIM $\uparrow$} & {MSE $\downarrow$} & {MAE $\downarrow$} & {PSNR $\uparrow$} & {SSIM $\uparrow$} \\
\midrule

\textsc{SimVP} & Baseline & 0.0828 & 0.1738 & 10.85 & 0.5607 & 0.0050 & 0.0335 & 22.99 & 0.8855 & 0.0016 & 0.0237 & 27.83 & 0.9951 \\
& + RAP    & {\cellcolor{bestcell} 0.0746} & {\cellcolor{bestcell} 0.1650} & {\cellcolor{bestcell} 11.31} & {\cellcolor{bestcell} 0.5818} & {\cellcolor{bestcell} 0.0006} & {\cellcolor{bestcell} 0.0123} & {\cellcolor{bestcell} 32.23} & {\cellcolor{bestcell} 0.9466} & {\cellcolor{bestcell} 0.0011} & {\cellcolor{bestcell} 0.0186} & {\cellcolor{bestcell} 29.72} & {\cellcolor{bestcell} 0.9968} \\
\addlinespace

\textsc{DiT} & Baseline & 0.0727 & 0.1674 & 11.42 & 0.5705 & 0.0693 & 0.2050 & 11.61 & 0.7441 & 0.0058 & 0.0475 & 22.37 & 0.9878 \\
& + RAP    & {\cellcolor{bestcell} 0.0677} & {\cellcolor{bestcell} 0.1593} & {\cellcolor{bestcell} 11.73} & {\cellcolor{bestcell} 0.5998} & {\cellcolor{bestcell} 0.0365} & {\cellcolor{bestcell} 0.0950} & {\cellcolor{bestcell} 14.40} & {\cellcolor{bestcell} 0.6103} & {\cellcolor{bestcell} 0.0005} & {\cellcolor{bestcell} 0.0117} & {\cellcolor{bestcell} 33.43} & {\cellcolor{bestcell} 0.9987} \\
\addlinespace

\textsc{Triton} & Baseline & 0.0522 & 0.1361 & 12.86 & 0.6451 & 0.0277 & 0.0631 & 15.59 & 0.8197 & 0.0002 & 0.0088 & 36.43 & 0.9991 \\
& + RAP    & {\cellcolor{bestcell} 0.0494} & {\cellcolor{bestcell} 0.1310} & {\cellcolor{bestcell} 13.10} & {\cellcolor{bestcell} 0.6599} & {\cellcolor{bestcell} 0.0221} & {\cellcolor{bestcell} 0.0484} & {\cellcolor{bestcell} 16.58} & {\cellcolor{bestcell} 0.8610} & {\cellcolor{bestcell} 0.0002} & {\cellcolor{bestcell} 0.0078} & {\cellcolor{bestcell} 37.07} & {\cellcolor{bestcell} 0.9993} \\
\addlinespace

\textsc{UNet} & Baseline & 0.1023 & 0.1982 & 9.94 & 0.5224 & 0.0744 & 0.1842 & 11.30 & 0.7220 & 0.0158 & 0.0704 & 18.00 & 0.9700 \\
& + RAP    & {\cellcolor{bestcell} 0.0994} & {\cellcolor{bestcell} 0.1957} & {\cellcolor{bestcell} 10.07} & {\cellcolor{bestcell} 0.5265} & {\cellcolor{bestcell} 0.0608} & {\cellcolor{bestcell} 0.1831} & {\cellcolor{bestcell} 12.18} & {\cellcolor{bestcell} 0.7506} & {\cellcolor{bestcell} 0.0125} & {\cellcolor{bestcell} 0.0616} & {\cellcolor{bestcell} 19.02} & {\cellcolor{bestcell} 0.9750} \\
\addlinespace

\textsc{CNO} & Baseline & 0.1237 & 0.2214 & 9.11 & 0.4833 & 0.0859 & 0.1388 & 10.68 & 0.6380 & 0.0372 & 0.1129 & 14.30 & 0.9427 \\
& + RAP    & {\cellcolor{bestcell} 0.1164} & {\cellcolor{bestcell} 0.2153} & {\cellcolor{bestcell} 9.38} & {\cellcolor{bestcell} 0.4954} & {\cellcolor{bestcell} 0.0351} & {\cellcolor{bestcell} 0.0982} & {\cellcolor{bestcell} 14.57} & {\cellcolor{bestcell} 0.7558} & {\cellcolor{bestcell} 0.0352} & {\cellcolor{bestcell} 0.1089} & {\cellcolor{bestcell} 14.53} & {\cellcolor{bestcell} 0.9461} \\
\addlinespace

\textsc{MGNO} & Baseline & 0.0693 & 0.1584 & 11.63 & 0.5944 & 0.0448 & 0.0958 & 13.50 & 0.6895 & {\cellcolor{bestcell} 0.1539} & {\cellcolor{bestcell} 0.2525} & {\cellcolor{bestcell} 8.13} & {\cellcolor{bestcell} 0.8332} \\
& + RAP    & {\cellcolor{bestcell} 0.0675} & {\cellcolor{bestcell} 0.1562} & {\cellcolor{bestcell} 11.74} & {\cellcolor{bestcell} 0.6024} & {\cellcolor{bestcell} 0.0086} & {\cellcolor{bestcell} 0.0430} & {\cellcolor{bestcell} 20.68} & {\cellcolor{bestcell} 0.8521} & 0.1542 & 0.2530 & 8.12 & 0.8318 \\
\addlinespace

\textsc{ResNet} & Baseline & 0.1163 & 0.2197 & 9.38 & 0.5142 & 0.0875 & 0.1821 & 10.60 & 0.6875 & 0.1736 & 0.2899 & 7.60 & 0.8900 \\
& + RAP    & {\cellcolor{bestcell} 0.1141} & {\cellcolor{bestcell} 0.2162} & {\cellcolor{bestcell} 9.46} & {\cellcolor{bestcell} 0.5221} & {\cellcolor{bestcell} 0.0529} & {\cellcolor{bestcell} 0.1655} & {\cellcolor{bestcell} 12.78} & {\cellcolor{bestcell} 0.7995} & {\cellcolor{bestcell} 0.1619} & {\cellcolor{bestcell} 0.2759} & {\cellcolor{bestcell} 7.91} & {\cellcolor{bestcell} 0.8941} \\
\bottomrule
\end{tabular}

\begin{tablenotes}[para,flushleft]
    \item[\textit{Note:}] Our RAP framework consistently improves performance. The best result in each model-benchmark comparison is highlighted with a \colorbox{bestcell}{light green background}. For MSE/MAE, lower is better ($\downarrow$), while for PSNR/SSIM, higher is better ($\uparrow$).
\end{tablenotes}
\end{threeparttable}
\end{table*}

\begin{figure*}[h!]
    \centering
    \includegraphics[width=1\textwidth]{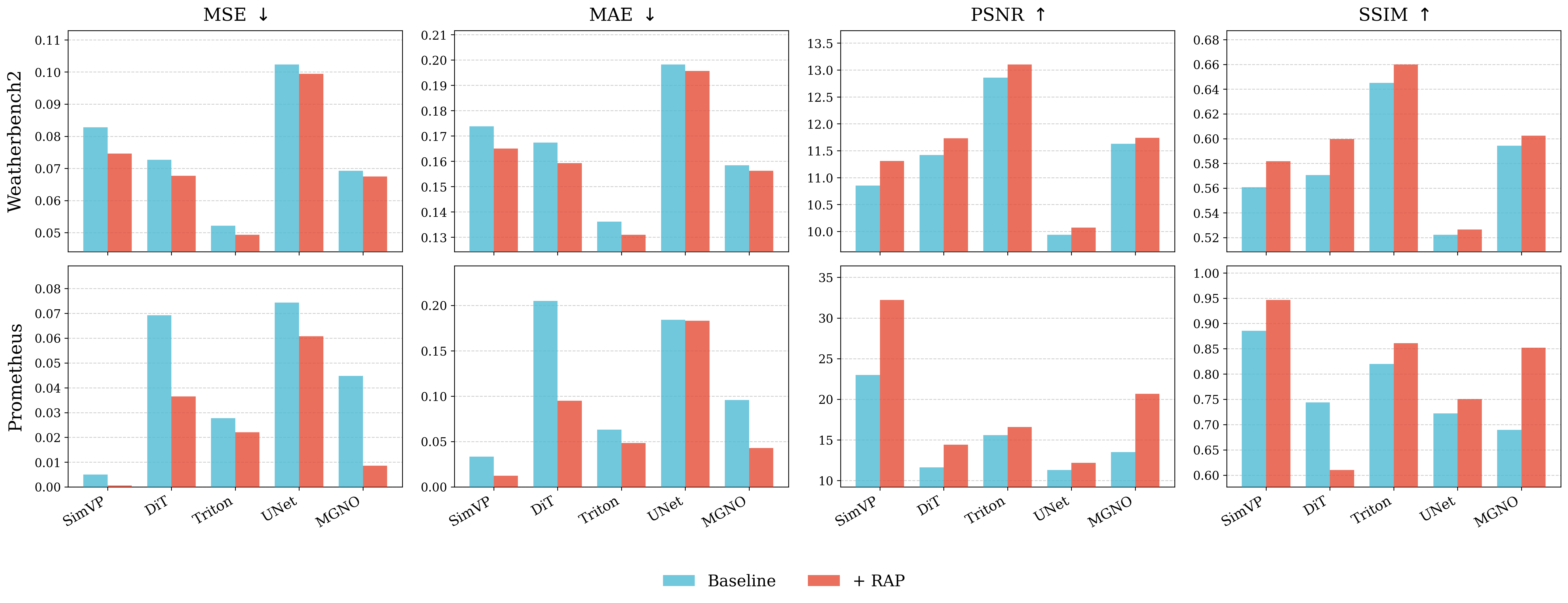}
\caption{
    \textbf{Quantitative comparison of baselines versus their RAP-enhanced counterparts across diverse benchmarks.}
        The figure provides a detailed performance comparison on the Weatherbench2 (top row) and Prometheus (bottom row) benchmarks, covering four key metrics: MSE ($\downarrow$), MAE ($\downarrow$), PSNR ($\uparrow$), and SSIM ($\uparrow$).
        It is evident that for all selected models, the RAP-enhanced versions (red) consistently and significantly outperform the original baseline models (blue) across all metrics. 
        This result strongly validates the effectiveness and generality of our RAP framework, highlighting its core contribution of leveraging historical analogs as explicit dynamical guidance to improve the accuracy of complex spatiotemporal prediction tasks.
}
\label{fig:relative_improvement}
\end{figure*}

\begin{figure*}[t]
    \centering
    \includegraphics[width=\textwidth]{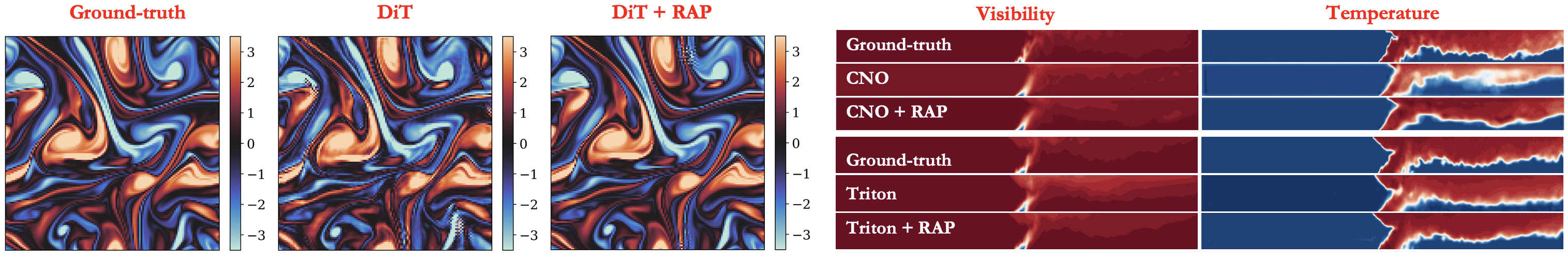}
    \caption{Visual comparison on Turbulence (left) and Prometheus (right). Unlike baseline models which blur results, RAP preserves sharp physical features like vortices and flame fronts.}
    \label{fig:qualitative_comparison}
\end{figure*}
\subsection{Main Quantitative Results}
\label{sec:main_results}

To answer \textbf{RQ1}, we first present a comprehensive quantitative comparison of our RAP framework against baseline models on the Weatherbench2 and Prometheus benchmarks. The detailed performance metrics are provided in Table~\ref{tab:main_results}, with a summary of the relative improvements visualized in Figure~\ref{fig:relative_improvement}. Our findings consistently demonstrate the effectiveness and generality of the RAP framework.

\paragraph{Consistent and Significant Improvements.}
As shown in Table~\ref{tab:main_results}, applying RAP (\textbf{+ RAP}) leads to a consistent and significant performance boost across all seven baseline models and on both diverse datasets. For instance, on the challenging Weatherbench2 task, RAP enhances the performance of all models, from the CNN-based UNet to the Transformer-based Triton. This universal improvement underscores that providing an explicit dynamical constraint through a historical analog is a fundamentally beneficial approach, regardless of the underlying model architecture.

\paragraph{Substantial Gains on Complex Dynamics.}
The advantages of RAP are particularly pronounced in tasks with highly non-linear and complex dynamics, such as the Prometheus fire spread simulation. Here, purely parametric models often struggle to capture intricate physical interactions. As highlighted in Figure~\ref{fig:relative_improvement}, RAP delivers remarkable gains for models such as SimVP, improving its MSE by an order of magnitude (from $0.0050$ to $0.0006$) and increasing its PSNR from $22.99$ to $32.23$. This corresponds to a relative improvement in the MSE of nearly 90\%, as seen in the bar chart. This indicates that the retrieved reference target provides crucial, high-fidelity information about the true physical evolution, effectively guiding the model away from physically implausible predictions and reducing error.

\paragraph{Enhancing State-of-the-Art Models.}
It is noteworthy that RAP enhances not only simpler models but also sophisticated, state-of-the-art architectures like Triton. On the Prometheus task, RAP improves Triton's MSE by \textbf{5.4\%} and its MAE by \textbf{3.7\%} (Figure~\ref{fig:relative_improvement}). While Triton is already optimized for physical system forecasting, it still benefits from the non-parametric guidance offered by a real-world evolutionary exemplar. This finding strongly supports our core hypothesis: combining the predictive power of parametric deep networks with the grounded truth of non-parametric historical experience is a powerful paradigm for building next-generation surrogate models. Similarly, for DiT, another strong Transformer-based model, RAP achieves a relative improvement of \textbf{6.9\%} in MSE and \textbf{5.1\%} in SSIM, demonstrating its broad applicability to powerful generative models as well.

In summary, our quantitative results provide compelling evidence that the RAP framework is a general, effective, and robust method for enhancing spatiotemporal prediction models. By leveraging historical analogs as explicit dynamical constraints, RAP consistently pushes the performance boundary across a wide spectrum of models and complex physical systems.

\subsection{Physical Fidelity and Stability}
\label{sec:fidelity}
To answer RQ2, we qualitatively evaluate the model's ability to generate physically plausible predictions, particularly during long-term rollouts where errors tend to accumulate. As shown in Figure~\ref{fig:qualitative_comparison}, purely parametric models often produce overly smooth or blurred artifacts that deviate from physical reality. For instance, in the 2D Turbulence task, the baseline DiT model fails to capture the fine-grained vortex structures. Similarly, in the Prometheus fire spread simulation, both the CNO and Triton models smooth out the sharp, chaotic flame front. In stark contrast, RAP-enhanced models consistently produce results with significantly higher physical fidelity. They successfully restored critical details, sharp vortices in turbulence and well-defined flame fronts in fire simulation, closely mirroring the ground truth. This compellingly demonstrates that by incorporating a real-world evolutionary exemplar, RAP serves as an effective dynamical constraint that guides the model toward physically plausible states and suppresses error divergence.

\subsubsection{On the Architectural Choice of the Reference Encoder}

To determine the optimal architecture for the reference encoder, we conducted a rigorous ablation study comparing a lightweight convolutional encoder (SimVP) against a more complex, attention-based DiT encoder. The results reveal a compelling insight: an {asymmetric encoder design} is superior. Across all tested backbones, the RAP framework equipped with the lightweight SimVP encoder consistently and significantly outperforms both the baseline and the version using the more complex DiT encoder. This is particularly evident with the state-of-the-art {Triton} backbone, where the SimVP encoder reduces the loss to {0.0494}, surpassing both the baseline (0.0522) and the DiT-enhanced version (0.0503).

The rationale for this finding lies in the specialized roles of the encoders. The reference target ($Y_{\text{ref}}$) is a complete, ground-truth trajectory, rich in direct structural patterns that a lightweight convolutional architecture like SimVP can efficiently extract. Conversely, a powerful Transformer is suboptimal for processing an input that already represents a physically valid "answer," as it may introduce unnecessary complexity. This insight underscores the modular strength of our framework, which allows for a flexible "mix-and-match" approach, tailoring encoder complexity to the nature of the input to build more efficient and specialized models.

\begin{table}[htbp]
\centering
\begin{threeparttable}
\caption{Ablation study on the architecture of the reference encoder. The experiment compares the baseline model, RAP with a DiT encoder, and RAP with a lightweight convolutional encoder (SimVP). The lightweight SimVP encoder consistently yields the best performance.}
\label{tab:ablation_encoder}
\small
\renewcommand{\arraystretch}{1.3}

\definecolor{headercolor}{gray}{0.92}
\definecolor{raprowcolor}{gray}{0.96}
\definecolor{bestcell}{RGB}{230, 245, 235} 

\sisetup{
    detect-weight,
    mode=text,
    table-number-alignment=center
}

\begin{tabularx}{\linewidth}{@{} l >{\raggedright\arraybackslash}X S[table-format=1.5] @{}}
\toprule
\rowcolor{headercolor}
\textbf{Backbone} & \textbf{Reference Encoder Configuration} & {\textbf{Loss} $\downarrow$} \\
\midrule

\textsc{SimVP} & Baseline (None) & 0.08285 \\
\rowcolor{raprowcolor}
& + RAP (DiT Encoder) & 0.07744 \\
\rowcolor{raprowcolor}
& + RAP (SimVP Encoder) & \cellcolor{bestcell}\bfseries 0.07459 \\
\midrule

\textsc{DiT} & Baseline (None) & 0.07275 \\
\rowcolor{raprowcolor}
& + RAP (DiT Encoder) & 0.06870 \\
\rowcolor{raprowcolor}
& + RAP (SimVP Encoder) & \cellcolor{bestcell}\bfseries 0.06770 \\
\midrule

\textsc{Triton} & Baseline (None) & 0.05224 \\
\rowcolor{raprowcolor}
& + RAP (DiT Encoder) & 0.05039 \\
\rowcolor{raprowcolor}
& + RAP (SimVP Encoder) & \cellcolor{bestcell}\bfseries 0.04944 \\
\midrule

\textsc{UNet} & Baseline (None) & 0.10234 \\
\rowcolor{raprowcolor}
& + RAP (DiT Encoder) & 0.09942 \\
\rowcolor{raprowcolor}
& + RAP (SimVP Encoder) & \cellcolor{bestcell}\bfseries 0.09936 \\
\bottomrule
\end{tabularx}

\begin{tablenotes}[para,flushleft]
    \item[\textit{Note:}] All models were trained for 500 epochs on the 1979--2017 dataset and evaluated on the 2018--2021 dataset. RAP methods utilize the `79-17 (interval 3)` set for retrieval. Rows with a \colorbox{raprowcolor}{light gray background} denote RAP-enhanced models. The best result for each backbone is highlighted in \colorbox{bestcell}{green}.
\end{tablenotes}
\end{threeparttable}
\end{table}

\begin{figure*}[h]
    \centering
    \includegraphics[width=\textwidth]{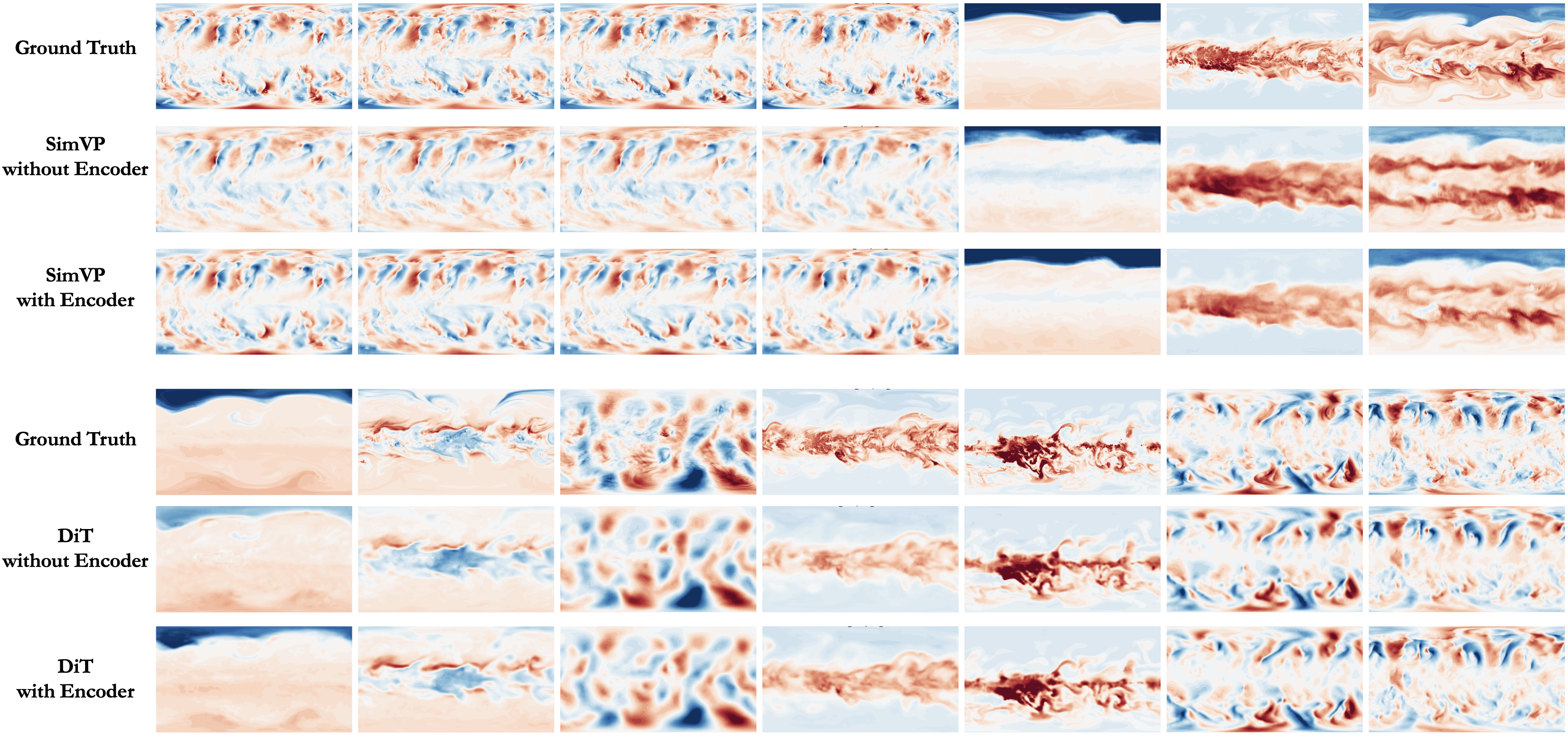} 
    \caption{Qualitative comparison on the Weatherbench dataset. The with Encodervariant corresponds to our full RAP framework, while without Encoder is the baseline model.}
    \label{fig:qualitative_fusion_necessity}
\end{figure*}

\subsection{Scalability and Efficiency with Large-Scale Models}

To evaluate the scalability and practical benefits of our framework ({RQ4}), we applied RAP to a {0.6B parameter Triton model}. We assess whether RAP can mitigate the performance loss from using a smaller training dataset by leveraging a larger retrieval database. The experimental setup and results are detailed in Table~\ref{tab:scalability}. The baseline model trained on the full dataset (1979-2017) sets a strong performance benchmark with a loss of {0.0946}. When the training data is reduced to a partial set (1993-2017), the baseline's performance degrades to {0.1003}, representing a performance gap of {0.0057}. Our key experiment involves training the model on the same partial dataset but augmenting it with RAP, using the full dataset as the retrieval database. This RAP-enhanced model achieves a loss of {0.0957}. This result demonstrates that by integrating historical knowledge through retrieval, our method recovers a substantial portion of the performance lost due to reduced training data. Specifically, the RAP framework closes the performance gap by {0.0046} (from 0.1003 to 0.0957), which accounts for approximately {81\%} of the total gap. This suggests that RAP can serve as a computationally efficient strategy for developing large models, where access to a large, static historical database can partially substitute for training on the entire dataset, thereby potentially reducing training costs.
\begin{table}[t]
\centering
\caption{
    \textbf{Scalability Analysis of RAP (0.6B Triton Model).} 
    Symbol key: (1) $\triangleright$=Partial data (93-17), $\square$=Full data (79-17), $\times$=No retrieval. 
    Our method ($\blacktriangleright$) with partial training but full retrieval achieves 99.2\% of full-dataset performance.
}
\label{tab:scalability}
\sisetup{
    table-number-alignment=center,
    table-figures-decimal=4,
    detect-weight=true,
    mode=text
}
\footnotesize
\begin{tabularx}{\linewidth}{@{}>{\raggedright}X c c c S[table-format=1.4]@{}}
\toprule
\textbf{Config} & \textbf{Train} & \textbf{Retrieval} & \textbf{Type} & \textbf{Loss} $\downarrow$ \\
\midrule
Baseline & $\triangleright$ & $\times$ & LB & 0.1003 \\
\rowcolor{gray!8}
RAP ($\blacktriangleright$) & $\triangleright$ & $\square$ & \textbf{Ours} & \textbf{0.0957} \\
Baseline & $\square$ & $\times$ & UB & 0.0945 \\
\bottomrule
\end{tabularx}
\vspace{-15pt}
\end{table}

\subsection{On the Necessity of the Dual-Stream Architecture}

To validate that the performance gains of RAP stem from its architectural design rather than merely from access to additional data, we conduct a critical ablation study. We investigate the efficacy of our proposed dual-stream fusion against a more naive integration strategy. The two compared methods are:

\begin{itemize}
    \setlength\itemsep{0em}
    \item[\ding{182}] \textbf{\textit{Dual-Stream Fusion (Ours)}}: Our proposed RAP architecture, which uses a dedicated reference encoder to extract dynamic patterns from the reference target ($Y_{\text{ref}}$) before fusing them at the feature level with the query representation.
    \item[\ding{183}] \textbf{\textit{Naive Concatenation}}: A simplified baseline where the reference target ($Y_{\text{ref}}$) is directly concatenated with the query input ($X_{\text{query}}$) along the channel dimension and fed into the original, unmodified backbone model.
\end{itemize}

\noindent\textbf{Quantitative Analysis.} The results, presented in Table~\ref{tab:ablation_fusion_necessity}, highlight the superiority of our dual-stream design. Across all backbones, our method consistently and significantly outperforms the naive concatenation approach. More revealingly, the naive strategy not only fails to improve upon the baseline but consistently degrades performance (e.g., Triton loss increases from 0.0522 for the baseline to 0.0639). We hypothesize that directly concatenating the reference, which has inherent spatiotemporal misalignments with the query, introduces conflicting signals that disrupt the model's learning process. This finding underscores a critical insight: \textit{how} the historical information is integrated is more crucial than its mere presence.

\noindent\textbf{Qualitative Analysis.} Visual evidence from the Weatherbench dataset, shown in Figure~\ref{fig:qualitative_fusion_necessity}, corroborates our quantitative findings. The baseline predictions (labeled without Encoder) exhibit significant blurring and a loss of high-frequency details, a typical artifact of numerical dissipation in long-term rollouts. In contrast, our full RAP framework (labeled with Encoder) successfully preserves sharp, physically plausible structures, such as fine-grained vortices and crisp weather fronts, maintaining high fidelity to the ground truth. This visual disparity confirms that our dual-stream architecture is essential for effectively parsing the dynamic guidance from the reference target and preventing the degradation of physical realism.

\begin{table}[h]
\centering
\begin{threeparttable} 
\caption{Ablation on the fusion strategy. Our proposed Dual-Stream Fusion consistently outperforms the Naive Concatenation baseline.}
\label{tab:ablation_fusion_necessity}
\small 
\renewcommand{\arraystretch}{1.25}
\definecolor{headercolor}{gray}{0.92}
\definecolor{bestcell}{RGB}{230, 245, 235}
\sisetup{
    detect-weight,
    mode=text,
    round-mode=places,
    round-precision=4,
    table-number-alignment=center
}
\begin{tabularx}{\linewidth}{@{}l X S[table-format=1.5]@{}}
\toprule
\rowcolor{headercolor}
\textbf{Backbone} & \textbf{Fusion Method} & {\textbf{Loss} $\downarrow$} \\
\midrule
\multirow{2}{*}{\textsc{SimVP}} 
 & Naive Concatenation & 0.0886 \\
 & Dual-Stream Fusion (Ours) & \cellcolor{bestcell}\bfseries 0.0746 \\
\addlinespace
\multirow{2}{*}{\textsc{DiT}} 
 & Naive Concatenation & 0.0843 \\
 & Dual-Stream Fusion (Ours) & \cellcolor{bestcell}\bfseries 0.0677 \\
\addlinespace
\multirow{2}{*}{\textsc{Triton}} 
 & Naive Concatenation & 0.0639 \\
 & Dual-Stream Fusion (Ours) & \cellcolor{bestcell}\bfseries 0.0494 \\
\addlinespace
\multirow{2}{*}{\textsc{UNet}} 
 & Naive Concatenation & 0.1132 \\
 & Dual-Stream Fusion (Ours) & \cellcolor{bestcell}\bfseries 0.0994 \\
\bottomrule
\end{tabularx}
\end{threeparttable}
\end{table}

\begin{table}[t] 
\centering
\begin{threeparttable}
\caption{Ablation on the impact of training data scale. RAP's benefits are complementary to the gains from a larger dataset.}
\label{tab:data_scale_ablation}
\footnotesize 
\renewcommand{\arraystretch}{1.2} 
\setlength{\tabcolsep}{3pt} 

\definecolor{headercolor}{gray}{0.92}
\definecolor{bestcell}{RGB}{230, 245, 235}

\sisetup{
    detect-weight,
    mode=text,
    table-number-alignment=center,
    round-mode=places,
    round-precision=5
}

\begin{tabularx}{1\columnwidth}{@{} l c >{\raggedright\arraybackslash}X c S[table-format=1.5] @{}}
\toprule
\rowcolor{headercolor}
\textbf{Backbone} & \textbf{Data} & \textbf{Method} & \textbf{Epochs} & {\textbf{Loss} $\downarrow$} \\
\midrule

\multirow{4}{*}{\textsc{SimVP}}
 & 93-17 & Base & 1000 & 0.08694 \\
 & 93-17 & + RAP & 500 & 0.08057 \\
 & 79-17 & Base & 500 & 0.08285 \\
 & 79-17 & + RAP & 500 & \cellcolor{bestcell}\bfseries 0.07459 \\
\midrule

\multirow{4}{*}{\textsc{DiT}}
 & 93-17 & Base & 1000 & 0.08139 \\
 & 93-17 & + RAP & 500 & 0.08039 \\
 & 79-17 & Base & 500 & 0.07275 \\
 & 79-17 & + RAP & 500 & \cellcolor{bestcell}\bfseries 0.06770 \\
\midrule

\multirow{4}{*}{\textsc{Triton}}
 & 93-17 & Base & 1000 & 0.05437 \\
 & 93-17 & + RAP & 500 & 0.05369 \\
 & 79-17 & Base & 500 & 0.05224 \\
 & 79-17 & + RAP & 500 & \cellcolor{bestcell}\bfseries 0.04944 \\
\midrule

\multirow{4}{*}{\textsc{UNet}}
 & 93-17 & Base & 1000 & 0.10929 \\
 & 93-17 & + RAP & 500 & 0.10686 \\
 & 79-17 & Base & 500 & 0.10234 \\
 & 79-17 & + RAP & 500 & \cellcolor{bestcell}\bfseries 0.09936 \\
\bottomrule
\end{tabularx}

\begin{tablenotes}[para,flushleft]
    \item[\textit{Note:}] "Base" denotes Baseline. The best loss for each backbone is highlighted in \colorbox{bestcell}{green}.
\end{tablenotes}
\end{threeparttable}
\end{table}

\begin{figure*}[t]
    \centering
    \includegraphics[width=0.99\textwidth]{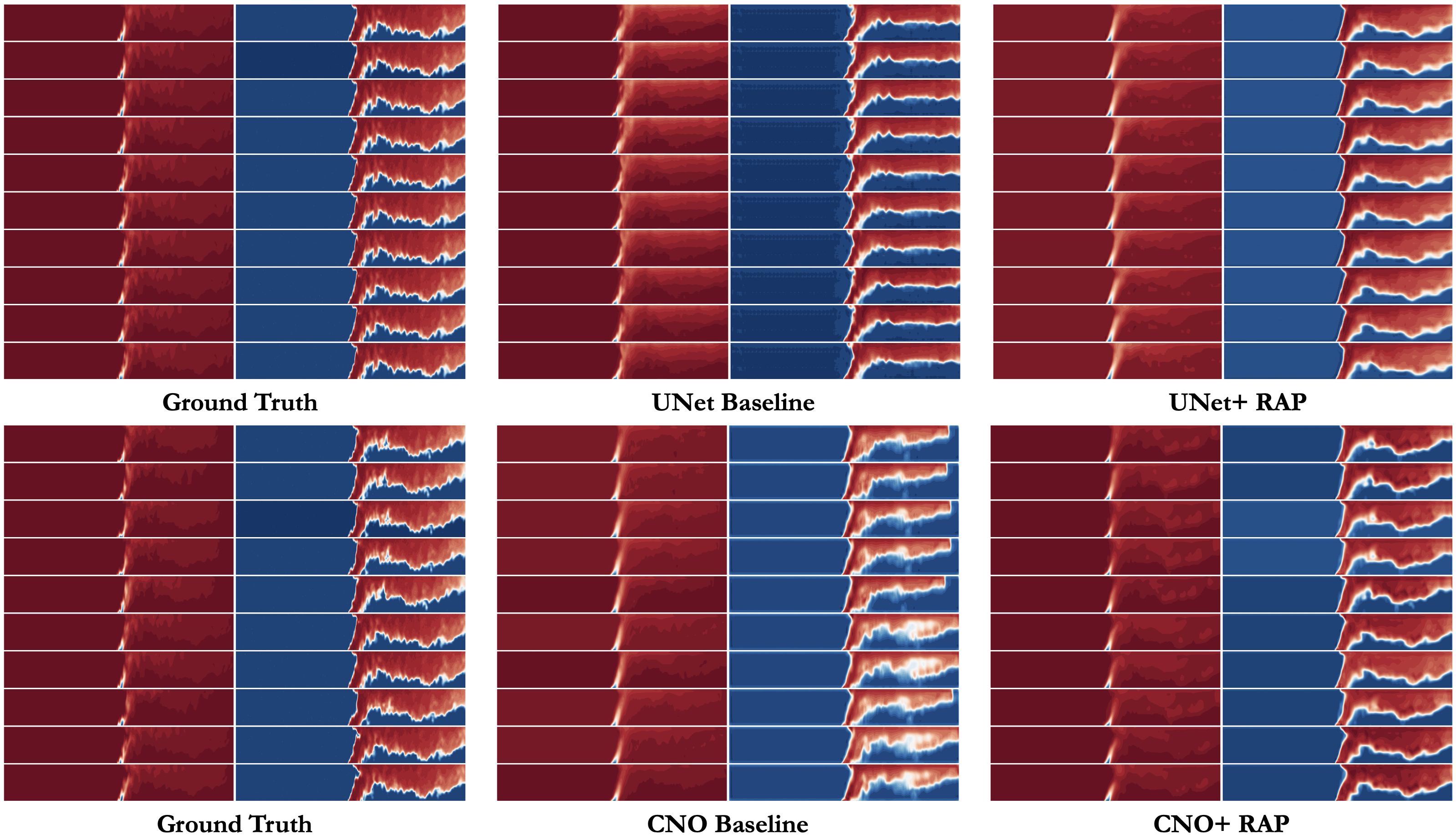}    
    \caption{
    Qualitative comparison of long-term predictions on the Prometheus fire spread simulation. 
    Each column block compares the \textbf{Ground Truth} (top row of each block), the \textbf{Baseline} model (middle), and the \textbf{RAP-enhanced} model (bottom). 
    The baseline models (UNet Baseline, CNO Baseline) exhibit significant smoothing artifacts and fail to capture the complex, sharp flame fronts characteristic of turbulent combustion. 
    In contrast, our RAP-enhanced models (UNet+ RAP, CNO+ RAP) successfully suppress numerical dissipation and restore high-fidelity, physically plausible details that closely match the Ground Truth.
    }
    \label{fig:prometheus_qualitative}
\end{figure*}
\subsection{Qualitative Evaluation of Physical Fidelity}

Our qualitative evaluation on the challenging Prometheus fire spread dataset visually confirms RAP's ability to enhance physical fidelity in highly nonlinear systems (Figure~\ref{fig:prometheus_qualitative}). Baseline models, such as UNet and CNO, exhibit severe smoothing artifacts, a visual manifestation of numerical dissipation where the sharp, intricate flame fronts are lost. In stark contrast, RAP-enhanced models successfully suppress this dissipation, restoring high-fidelity flame fronts that are topologically consistent with the ground truth, complete with their intricate curls and sharp tips. This compellingly demonstrates that by introducing a real-world evolutionary exemplar as dynamic guidance, the RAP framework imposes a powerful physical constraint that effectively anchors predictions onto a more physically credible trajectory.

\subsection{Impact of Training Data Scale}

To investigate the interplay between RAP and data scale, we compared models trained on a partial (93-17) and the full (79-17) dataset. The results in Table~\ref{tab:data_scale_ablation} compellingly show that RAP's benefits are \textbf{complementary} to data scaling. For instance, while expanding the training data improves the Triton baseline's loss from 0.05437 to 0.05224, applying RAP to this already stronger baseline yields an additional significant reduction to \textbf{0.04944}. This demonstrates that RAP offers a unique form of dynamical guidance that is not fully captured even when training on a larger dataset, highlighting its synergistic value.

\section{Conclusion}

The Retrieval-Augmented Prediction (RAP) framework significantly enhances spatiotemporal forecasting capabilities by introducing a novel paradigm. The core idea of this approach is to leverage historical analogs past, true evolutionary exemplars that are most similar to the current state as an explicit dynamical constraint. This guidance steers deep learning models toward generating predictions that are more consistent with physical laws. Across various complex physical systems, including weather forecasting, turbulence modeling, and fire spread simulation, our experiments consistently demonstrate that this framework improves the accuracy of state-of-the-art models. Furthermore, it effectively suppresses the error accumulation and detail degradation commonly observed in long-term autoregressive rollouts, leading to a qualitative leap in the physical fidelity of the predictions.

As a general and scalable strategy, the RAP framework presents a promising pathway toward building the next generation of high-fidelity surrogate models for complex physical systems. It synergizes the powerful fitting capabilities of parametric, data-driven models with the grounded truth of non-parametric historical data, affirming the immense value of the classic learning from history philosophy in modern scientific computing. Future work will explore more advanced retrieval techniques and dynamic weighting mechanisms to unlock its potential across a broader spectrum of scientific and engineering domains.

\section{Limitations and Future Directions}

Despite its demonstrated effectiveness, the current RAP framework defines clear avenues for future work. Our reliance on a pixel-wise MSE for retrieval, while serving as a robust baseline, may not fully capture the dynamical similarity between states, especially in the presence of spatial shifts. Consequently, we plan to explore more sophisticated, physically-aware similarity metrics, such as those based on optimal transport or learned representations. Furthermore, the framework currently leverages only the single best analog; future iterations will investigate techniques for retrieving and fusing multiple historical analogs to provide a more comprehensive form of dynamic guidance. Finally, we aim to develop adaptive gating mechanisms that allow the model to dynamically weigh the retrieved guidance against its own parametric prediction, potentially based on the quality or relevance of the match.

\clearpage

\section*{Al-Generated Content Acknowledgement}
We used Gemini2.5 solely to polish the grammar of the main text. Apart from this, no AI-generated content (AIGC) was used.

\balance
\bibliographystyle{plain}
\bibliography{my_bib}

\end{document}